# A Representation of Uncertainty to Aid Insight into Decision Models

Holly B. Jimison

Medical Computer Science
Medical School Office Building Room X215
Stanford University, Stanford, California   94305
(415)723-2956      Jimison@Sumex-Aim.Edu

## Abstract:

Many real world models can be characterized as *weak*, meaning that there is significant uncertainty in both the data input and inferences. This lack of determinism makes it especially difficult for users of computer decision aids to understand and have confidence in the models. This paper presents a representation for uncertainty and utilities that serves as a framework for graphical summary and computer-generated explanation of decision models. The implementation described is a computer decision aid designed to enhance the clinician-patient consultation process for patients with suspected angina (chest pain due to lack of blood flow to the heart muscle). The generic angina model is represented as a Bayesian decision network, where the probabilities and utilities are treated as random variables with probability distributions on their range of possible values. The initial distributions represent information on all patients with anginal symptoms, and the approach allows for rapid customization to more patient-specific distributions. The framework provides a metric for judging the importance of each variable in the model dynamically.

## Introduction

It is often difficult to give an intuitive summary and explanation of a complicated decision problem. Yet, if a computer decision-support system is to be successful, it is important that it provide the user with insight into the decision model and a justification for the resulting recommendations. Many decisions must be made in a timely manner with incomplete and sometimes unreliable information. Computer decision aids can provide valuable assistance for these types of problems, however, it is important that a system provide an intuitive overview and explanation for the resulting recommendation.

In our design of a computer system to augment the clinician-patient consultation process we chose an approach that takes advantage of a decision-analytic framework but optimizes efficiency and user understanding. The clinical domain of angina was chosen as a testbed for this methodology because of the significant uncertainty associated with the choice of possible tests and treatments. Also, patient preferences and lifestyles were often found to be important variables in this domain, as in many others, and these factors are difficult for clinicians to incorporate into decision making when dealing

189

with what is already a complex medical model. The system was envisioned as a tool to augment the cardiologist - angina patient consultation process. Information must pass from the patient to the cardiologist as well as from the cardiologist to the patient. The goals for the computer system were to improve the quality of communication in this setting and to improve both the clinician's and the patient's understanding of what aspects of the complicated decision model are important for this particular patient.Because of this perspective, the system is referred to as the *Angina Communication Tool*.

The first entrance into the angina model comes in distinguishing nonspecific or nonischemic chest pain from anginal pain. The nonanginal patients are typically sent home, possibly with trial medications. The explanation given by the system here is fairly uniform and doesn't require a complicated model. In contrast, the decision of whether or not to do a moderately invasive procedure like angiography in anticipation of a possible bypass surgery is much more difficult to make patient-specific and to explain intuitively. It is this latter type of decision problem that is more challenging and interesting to summarize and explain. It is this aspect of the system that will be the focus of this paper.

## Knowledge Representation

The model used in the *Angina Communication Tool* to represent the many decisions encountered when working up a patient with chest pain symptoms is represented as a Bayesian Decision Network (also known as Influence Diagrams[2,3]).
This formalism is similar to that of Causal Networks[4,5,6] and Belief Networks[7]. However, in order to provide a metric for judging the sensitivity, importance and the degree of accuracy required for each variable, the model needs to be framed in terms of a decision, where the possible treatment alternatives and the utilities of the possible outcomes are explicitly modeled. There is an equivalent decision tree representation of the model, but the graphical and visual complexity of decision trees explodes exponentially with the addition of new variables while the number of new nodes in a network grows only linearly. This in itself makes the network representation more useful for both the model builder and the end user.

The representation of uncertainty is an important and integral part of these network models. The nodes represent uncertain model variables and the directed arcs between them show conditional probabilistic dependence. For example, in Figure 1 below the probability of B depends on the value of A. In other words $P(B|A) \neq P(B)$. Also, this model is necessarily an acyclic graph, where in this example knowledge about the state of E is fully modeled and characterized by knowledge of C and F.

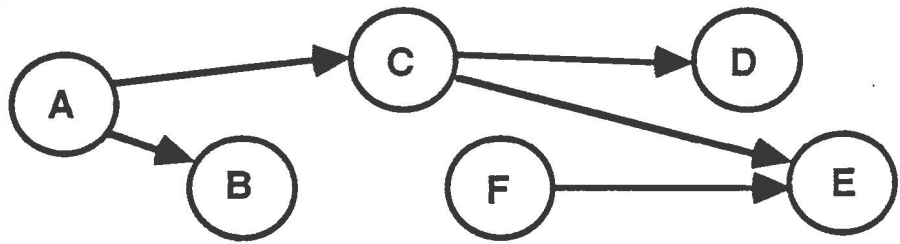

Figure 1. Bayesian Decision Network

To augment the basic framework of the network representation described above, the probabilities and utilities in the *Angina Communication Tool* are represented as random variables. Each random variable has a probability distribution over its set of possible values. These distributions allow the dynamic calculation of expected value of information for each variable and dynamic sensitivity analysis for the overall model. These features provide formal metrics for judging the importance of variables in the model, ordering patient assessment questions, structuring the graphics, and generating text explanation.

## Customizing a Generic Model

In order to streamline the task of modeling a specific patient's decision in real time (during a consultation), the *Angina Communication Tool* starts with a generic model of the domain. The general structure, consisting of alternative tests and treatments, chance events, and possible outcomes is applicable to all patients presenting with chest pain. The prior distributions for the probabilities and utilities of the model represent the population of all patients encountered in this domain. The "typical" patient is described using the mean of each distribution and what is usually prescribed is defined by the test or treatment with the highest mean expected utility. These distributions are refined and narrowed using information from patient assessment questions. These questions serve as predictors that assign patients to more homogeneous subgroups with more specific distributions. The figures below demonstrate how this occurs for a probability random variable and a utility random variable.

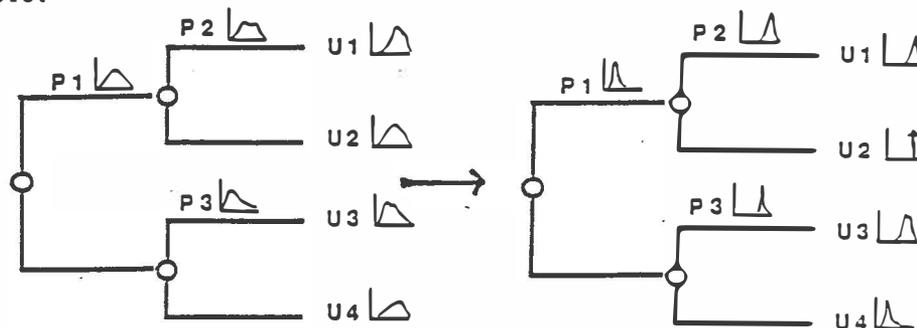

Figure 2a    Generic Model --> Patient-Specific Model

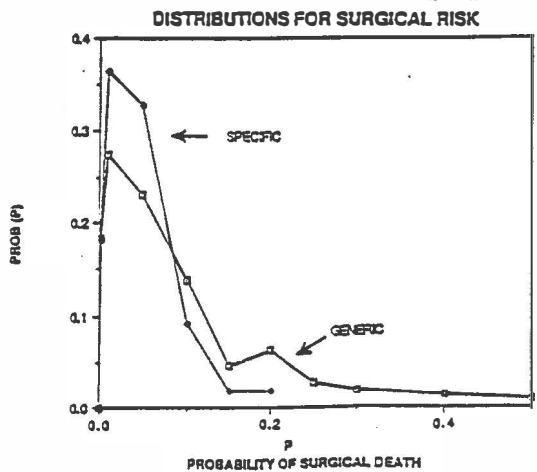

Figure 2b.

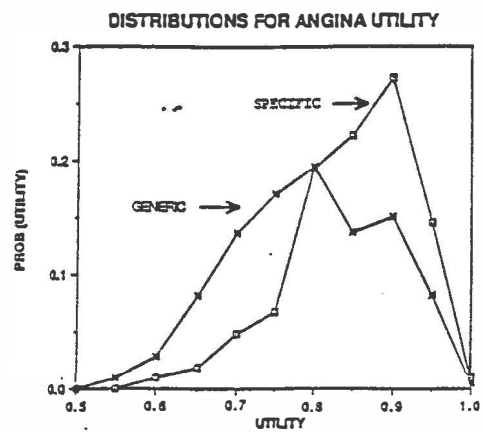

Figure 2c.

Figure 2a shows a small section of a decision tree with the probabilities and utilities represented as random variables with their associated distributions. The top section is part of the generic model, where the wide distributions are the prior probability distributions reflecting information from the whole patient population. With data from the patient assessment questions these distributions become more patient-specific, as is shown in the bottom section of Figure 2a. The new distributions come from data on more homogeneous subgroups with which the patient in consultation can be associated. Figure 2b shows how the prior distribution for the probability of surgical death becomes more specified after learning the patient's age, sex, and other diagnosed diseases. For the random variable of the utility of the quality of life with angina pain, Figure 2c shows that the wide prior distribution of utility values is narrowed after the system learns whether the patient has a basically sedentary or active lifestyle required for work and outside interests. This Figure shows the resulting distribution for a sedentary patient that does not devalue living with occasional angina pain as much as the group of more active patients.

One advantage to working with distributions as opposed to point probabilities and utilities is that the questions can be designed to be more clear and easy for a patient to answer. This is in contrast to traditional assessment techniques where the patient must understand probability, the possible medical outcomes, and the notion of choosing between hypothetical lotteries and/or hypothetical time trade-off questions. Currently, data for contructing the distributions on probability variables exists for some of the relevant subgroups of patients, but with a few exceptions[7,8] the population data necessary for utility models does not exist for clinical applications. Ideally, the data would be carefully collected from patients familiar with the medical outcome of interest and also educated with regard to the assessment techniques. In the meanwhile, subjective distributions are assessed from an expert during system development. Experts can either give subjective frequency distributions or simply provide parameters for assumed parametric distributions. The goal of the approach is to trade off the added complexity of system design for user interface and decision quality benefits derived during real time use. The distribution means will always be used to decide among treatment and test alternatives at any point in the consultation, but it will be shown that the variances of the distributions are important in directing the interaction with the patient and in placing appropriate emphasis on the most important variables in the model.

The design phase of a system that uses this type of representation includes the selection of possible patient assessment questions that most efficiently narrow the distributions for the variables in the model and separate the means for the expected utilities of the alternatives. In real-time use the assessment questions are ranked by *expected value of information*. The assessment questions also have a cost associated with them, so the modeling of this cost must heuristically put in the same units or scale as the *expected value of information*. In the Angina Communication Tool the cost of the assessment questions are assigned based on the estimated time and difficulty associated with the question. Questions on age, sex, and test results already noted in the chart are basically free of cost, while a set of hypothetical standard gamble questions used to obtain a point utility would be much more costly. Functional questions such as number of blocks walked per day would fall somewhere in between. This metric comparing expected value of information of a question versus its cost serves to dynamically order the assessment questions and provides a suggested stopping rule that notes when the cost of asking a question exceeds its expected value of information. This approach optimizes the efficiency of patient-specific modeling. The major part of the model building is done during system development in the creation and encoding of a generic model. The real time use in the clinician-patient consultation simply involves customizing the generic model to the patient at hand. The ordering of the assessment questions then optimizes this customization by focusing on the most important variables for that patient and minimizing the time and complexity of assessment by letting the user know when the decision is sufficiently robust and dominant.

## Graphical Interface

A major focus and design goal for the *Angina Communication Tool* was to provide the clinician and patient with an intuitive understanding the factors most critical to deciding among treatment alternatives for the patient and to provide feedback on the progress of the assessment procedure. To do this effectively it was necessary to keep the general graphic display as simple as possible without hiding important information. The overall display has four major windows:

    A program control and *Help* window (menu-driven with a mouse input device),

    A window for displaying information associated with any variable node (these are encoded as objects with attributes that can be selected by buttoning the graphical representation with a mouse input device);

    A window that continuously gives a concise graphical overview of the patient's decision model, and

    A text window that prompts the clinician with assessment questions and when requested provides a computer-generated summary and explanation of the recommendation.

The graphics in the third window mentioned above consists of two sections. The first is a very simple histogram representation of the distributions on expected utilities for each of the relevant alternatives. The example shown in Figure 3 considers the decision between treating the patient with medications or performing cardiac bypass surgery. This is the appropriate decision context for a patient that has already been classified as high risk and potentially suitable for surgery. If angiography had not yet been done the appropriate decision context would involve the choice of performing or not performing angiography, and this decision would anticipate the later decision between bypass surgery or treating with medications.

The importance of this display is that it gets updated with each new piece of information about the patient. The expected utilities and the associated error bars are computed using Monte Carlo simulation. To do this, a point is randomly chosen from each cumulative distribution, thereby selecting representative samples from the probability distributions for each variable. Sampling from each distribution essentially provides one instantiation of the overall model, as if it were the description of a particular patient. The expected utility for each alternative is calculated for this instantiation and these points contribute to the histogram distribution on expected utilities for

193

repeated instantiations. The overlap of the expected utility distributions for the alternatives gives the user of the system a feel for how dominant the decision is, given what is known about the patient. This is a characterization of overall model sensitivity, and this information is displayed for the user as a dynamically changing histogram of mean expected utilities for the relevant alternatives as shown in Figure 3. The error bars on the graph represent +/- either one or two standard deviations of expected utility (choice toggled at the users discretion). The feedback showing changes in mean expected utilities as well as the overlap between alternatives provides the user with a sense of how the assessment is progressing and how robust the decision would be at that point in time.

The second graphical section in the model overview window is a dynamically changing representation of the patient's Bayesian decision network. This graphical view is simplified and made intuitive by sizing and pruning nodes from the display based on the variable's sensitivity with respect to the decision. The larger nodes have more influence on the decision and those variables that would not change the resulting decision can be safely hidden from view. If desired, the user can temporarily reexpand the graphical representation of the model. Nodes of variables with unusual values(when compared with the prior distribution are given a different pattern fill.

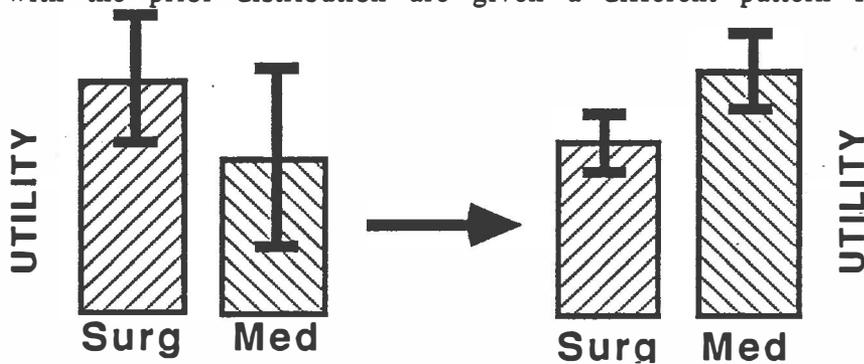

Figure 3. Dynamic Display of Overall Model Uncertainty

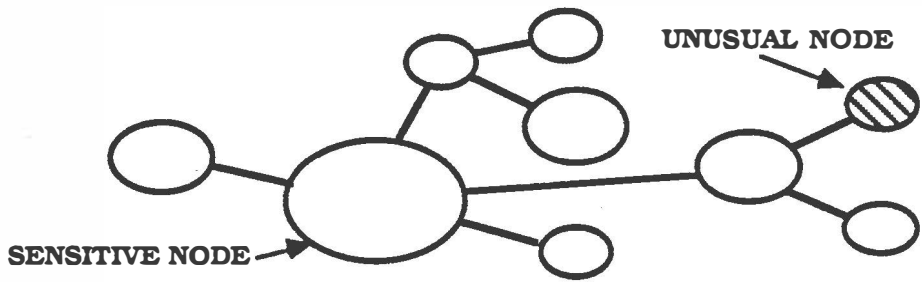

Figure 4. Nodes Emphasized According to Sensitivity

These two forms of graphical model overviews serve to aid the clinician and patient in focusing on what is currently most important for the patient's decision, while also giving a sense of how the model refinement is progressing. The purpose of these system features is to enhance the information transfer between the clinician and patient, both in understanding and efficiency.

## Text Explanation

The model and uncertainty representations for the *Angina Communication Tool* were chosen to provide a framework for efficient computer-generated explanation. The basic idea behind the explanation methodology was to assume an understanding of the generic model on the part of the clinician. An optional summary of the generic model is available on request. To keep the explanation of the patient's model concise, differences between the two are emphasized. Also, using the same metric of expected value of information and sensitivity as described previously, the variables can be ranked according to importance for explanation.

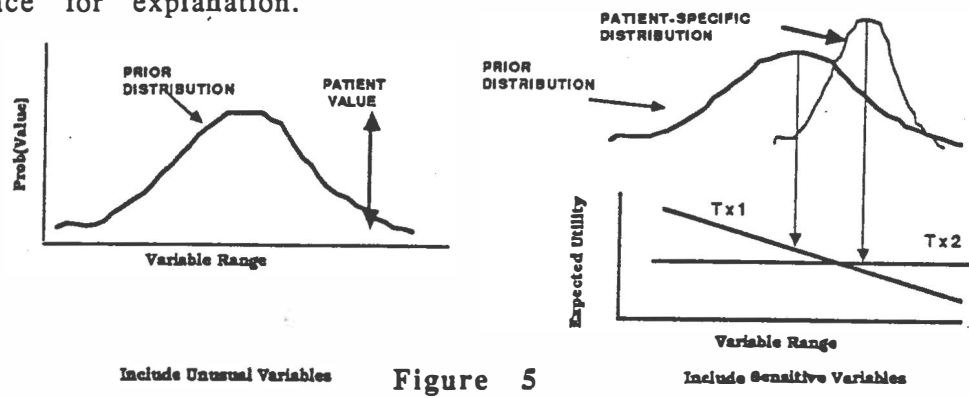

Figure 5

Figure 5a shows an unusual patient value compared to the prior distribution. If this variable were age, for example, the explanation would mention the patient's extreme advanced age and also mention the factors dependent upon age (eg surgical risk). In Figure 5b we note that the mean of the patient-specific distribution is not very different from the prior generic distribution, however, this change critically affects the choice between treatments 1 and 2. Thus, it gets flagged for explanation of the patient-specific model. An example of this is comes from the distribution data for the utility of angina pain, as shown previously. In both cases the patient-specific distribution is referenced to the prior generic distribution to measure its importance for explanation. Once again the methodology serves to focus the attention and communication on aspects of the decision model that are most relevant for the patient.

## Conclusions

A method for representing the generic knowledge about test and treatment decisions for the medical domain of angina has been described. The probabilities and utilities in the model are represented as random variables with distributions on their sets of possible values. The prior distributions define a generic model which is made patient-specific through an assessment process that dynamically creates an optimal ordering for questions according to a metric based on each question's expected value of information. The approach allows for the rapid customization of the model while providing a metric for judging the importance of each variable in the model. The graphical interface uses this information to display interactively a concise representation of the overall model and its associated uncertainty. Text explanation and summary of the patient-specific model is referenced to an assumed understanding of what is normally done for the typical patient, as defined by the generic model. The explanation emphasizes components of the patient's model that are sensitive and/or deviate from what is typically observed. These techniques serve to keep the explanation of the decision model concise, allowing the clinician and patient to focus on the issues that are most important for that individual patient.

## Acknowledgements

Harold C. Sox, Jr. and Charles Dennis have made essential contributions to the clinical aspects of this project, while M. Pavel has provided invaluable advice in the areas of measurement theory and user interface. This work was supported in part by a traineeship from the National Library of Medicine, with computing facilities provided by the SUMEX-AIM resource under the NIH Grant RR-00785.